\documentclass{article} 
\usepackage{iclr2021_conference,times}

\usepackage{amsmath,amsfonts,bm}

\def\eqref#1{equation~\ref{#1}}

\def\1{\bm{1}}

\DeclareMathAlphabet{\mathsfit}{\encodingdefault}{\sfdefault}{m}{sl}
\SetMathAlphabet{\mathsfit}{bold}{\encodingdefault}{\sfdefault}{bx}{n}

\usepackage{hyperref}
\usepackage{url}
\usepackage{graphicx}
\usepackage{chngcntr}

\title{Studying the Consistency and Composability of Lottery Ticket Pruning Masks}

\author{Rajiv Movva \\
MIT EECS\\
\texttt{\small rmovva@mit.edu} \\\And 
Jonathan Frankle \\
MIT CSAIL\\
\texttt{\small jfrankle@csail.mit.edu}\\\And 
Michael Carbin \\
MIT CSAIL\\
\texttt{\small mcarbin@csail.mit.edu}
}

\iclrfinalcopy 
\begin{document}

\maketitle

\begin{abstract}
Magnitude pruning is a common, effective technique to identify sparse subnetworks at little cost to accuracy. In this work, we ask whether a particular architecture's accuracy-sparsity tradeoff can be improved by combining pruning information across multiple runs of training. From a shared ResNet-20 initialization, we train several network copies (\textit{siblings}) to completion using different SGD data orders on CIFAR-10. While the siblings' pruning masks are naively not much more similar than chance, starting sibling training after a few epochs of shared pretraining significantly increases pruning overlap. We then choose a subnetwork by either (1) taking all weights that survive pruning in any sibling (mask union), or (2) taking only the weights that survive pruning across all siblings (mask intersection). The resulting subnetwork is retrained. Strikingly, we find that union and intersection masks perform very similarly. Both methods match the accuracy-sparsity tradeoffs of the one-shot magnitude pruning baseline, even when we combine masks from up to $k = 10$ siblings. 
\end{abstract}

\section{Introduction}

Recent work on the lottery ticket hypothesis has shown that sparse subnetworks within a dense architecture can be retrained to similar accuracies as their parent network \citep{frankle_lottery_2019}. These subnetworks are determined via magnitude pruning (MP) after training: the lowest magnitude weights are masked to zero, and the remaining weights are retained. However, little work has explored the consistency of masks sourced by MP: is the same subnetwork identified when there is different training noise? \citet{paganini_bespoke_2020} show that, for a given architecture and initialization, there exist multiple, dissimilar masks that retain accuracy. These distinct masks can be identified by using different training datasets or pruning methods. Here, we study how merely changing the sample of SGD noise affects magnitude pruning masks, and explore the shared structure among such masks in an effort to identify a subnetwork with improved accuracy-sparsity tradeoffs.

Specifically, our work investigates if the pruning masks of two networks \textit{compose}: that is, if the two masks can be combined, via union or intersection, to yield a single mask that reaches improved accuracy at a given sparsity. We create several copies of ResNet-20, termed \textit{siblings}, after some amount of pretraining. Each sibling is trained to completion using different samples of SGD noise, and we identify each sibling's mask by magnitude pruning. One hypothesis is that only weights pruned by all siblings can be confidently removed without affecting accuracy. Another hypothesis is that it is only necessary to retain the weights that no sibling prunes. Accordingly, we evaluate two strategies for combining masks:

\begin{enumerate}
    \item \textit{Mask union}: retain all weights that survive pruning in at least one sibling.
    \item \textit{Mask intersection}: retain only the weights that survive pruning across all siblings.
\end{enumerate}

To test each hypothesis, we compare the accuracy-sparsity tradeoffs of these two strategies to a standard one-shot pruning baseline \citep{frankle_linear_2020}. For both the union and intersection masks, we test for three possible outcomes: that the masks yield (A) better, (B) worse, or (C) the same accuracy-sparsity tradeoffs as the baseline. If mask union has the best accuracy-sparsity tradeoffs, we would conclude that pruning weights conservatively, only when siblings agree, optimally retains accuracy despite lower sparsity. Meanwhile, if mask intersection has the best tradeoffs, we'd conclude that keeping only the consistently unpruned weights is able to maintain accuracy while achieving higher sparsities. Finally, both strategies losing to the baseline would suggest that the masks are not complementary enough to be composable with union or intersection.

\textbf{Overview.} To assess the feasibility of composing masks, we start by studying the consistency of pruning masks across different training runs. We test whether sibling networks prune similar weights, when the only difference in their training process is their SGD noise (Section 2). We then ask how the overlap of pruned weights for a set of masks relates to their composability: that is, the accuracy-sparsity tradeoffs of the union or intersection mask (Section 3). We perform experiments using ResNet-20, trained and evaluated on CIFAR-10.


\textbf{Findings.} We provide evidence in support of the following, for ResNet-20:
\begin{itemize}
    \item Despite a shared initialization, sibling networks trained using different SGD orders prune largely dissimilar sets of weights (close to random chance overlap).
    
    \item Pretraining the network before creating siblings notably increases pruning overlap.
    
    \item Strikingly, for siblings that have been pretrained, taking either the union or the intersection of their masks yields similar accuracy-sparsity tradeoffs to the one-shot pruning baseline. This result holds even when we combine up to $k=10$ sibling masks.
\end{itemize}

\section{How similar are lottery ticket pruning masks?}

For a given network initialization, it is unclear whether magnitude pruning masks remain consistent in response to noise in the training process. Here, we study the similarity of masks sourced from different samples of SGD noise, motivated by the hypothesis that masks with substantial pruning overlap may be more composable, \textit{i.e.} perform better when we take their union or intersection.

\textbf{Pruning methodology.} Consider a neural network $\mathcal{N} = f(x; \theta)$, with inputs $x$ and parameters $\theta \in \mathbb{R}^d$. A \textit{pruning mask} is a binary vector $m \in \{0, 1\}^d$ that yields a corresponding subnetwork, $f(x; m \odot \theta)$. That is, the subnetwork is a copy of $f(x; \theta)$ with some weights zeroed out, or ``pruned.'' To determine $m$, we use magnitude pruning after training the dense network to completion \citep{han_learning_2015}. That is, for a chosen target sparsity $s$, we prune the $(s\cdot d)$ weights with the lowest magnitudes: $m[i] = 0$ for pruned weights, while $m[j] = 1$ for the others. We prune to the target sparsity in a single pass (\textit{one-shot} pruning), in contrast to pruning iteratively.

\textbf{Training sibling networks.} Given a network $\mathcal{N}$ with weights $\theta$, we create $k$ copies of $\mathcal{N}$, called sibling networks. The sibling networks' starting weights are set to either $\mathcal{N}$'s random initialization, $\theta = \theta_0$, or the result of pretraining $\mathcal{N}$ for $t$ iterations, $\theta = \theta_t$. We train each sibling to completion with different samples of SGD noise (\textit{i.e.}, different data orders and augmentations), producing trained weights $\theta_T^1, \cdots, \theta_T^k$. We use the ResNet-20 architecture and the CIFAR-10 dataset for all experiments, using state-of-the-art hyperparameters from \citet{frankle_linear_2020}.


\textbf{Comparing networks.} Next, we evaluate how consistently weights are pruned. We prune each sibling to a sparsity $s$, yielding masks $m^1, \cdots, m^k$ each with $(s \cdot d)$ weights masked to zero ($\theta$ has $d$ total weights). We compute the pruning overlap ratio, \textit{i.e.} the fraction of pruned weights that all siblings prune: $$\text{overlap ratio} = \left|(m^1 = 0) \cap \cdots \cap (m^k = 0)\right| / (s \cdot d).$$ We also compute the overlap ratio expected by chance, for $k$ random masks at sparsity $s$: $\displaystyle (s^k \cdot~d) / (s \cdot~d) = s^{k-1}$.

\begin{figure}[!ht]
\begin{center}
    \includegraphics[width=\textwidth]{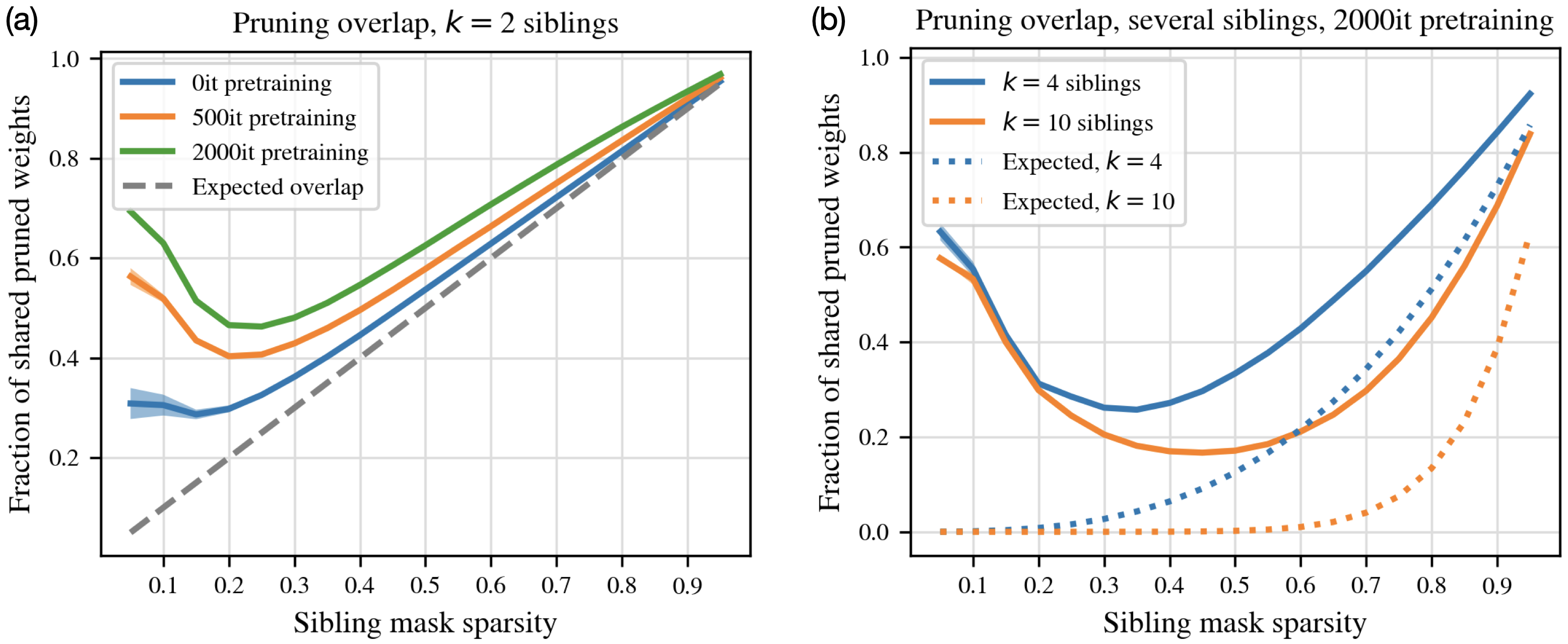}
    \caption{Pruning overlap, \textit{i.e.} the fraction of weights that each sibling prunes that all other siblings also prune. (a) $k=2$ siblings, variable pretraining; (b) 2000it pretraining, $k=4$ or $k=10$ siblings.}
    \label{fig:overlap}
\end{center}
\end{figure}

\textbf{Results.} At all sparsities, the pruning overlap ratio for a pair of siblings exceeds random chance, even without any shared pretraining (Figure \ref{fig:overlap}a, blue line). This is especially true at low sparsities: when each sibling is pruned to $s \le 20\%$, they share at least $30\%$ of their pruned weights (compared to the $20\%$ by chance). However, at $s > 20\%$, the overlap for siblings without pretraining tends close to that expected by random chance.

Shared pretraining prior to branching the siblings significantly increases pruning overlap. Even 500 iterations of pretraining ($<1\%$ of full training) yields significantly more overlap than siblings with no pretraining, especially up to moderate sparsities. With 2000 iterations of pretraining, pruning overlap substantially exceeds random chance at all sparsities.

We further study the consistency of pruned weights by extending our overlap analysis to several siblings (Figure \ref{fig:overlap}b). We compute masks after training $k=4$ or $k=10$ siblings with 2000 iterations of pretraining. Strikingly, both $k=4$ (blue) and $k=10$ (orange) siblings dramatically exceed the random overlaps expected by chance. For $s \le 60\%$, the $k=10$ curve is higher than the random overlap expected for four siblings, and is higher than the ten sibling expected overlap curve by multiple orders of magnitude. These experiments present additional evidence that certain pruned weights are indeed robust to SGD noise. Together, our results show that (1) pretraining siblings significantly increases mask overlap, and (2) compared to random chance, pretrained siblings prune remarkably consistent sets of weights.

\section{Combining information across pruning masks}

In Section 2, we showed that sibling masks prune more of the same weights than expected by chance, and that pretraining can significantly increase this overlap. In this section, we operationalize that finding: when siblings prune similar weights, we show that their masks can be composed via union or intersection, yielding a single mask that retains task accuracy.

\textbf{Mask composition.} We propose two strategies to compose masks: (1) mask union (keeping weights if they are in any mask, or equivalently pruning only the weights pruned by all masks), and (2) mask intersection (keeping weights only if they are present in all masks). We evaluate the composed mask $\mathcal{M}$ as follows: rewind the network weights $\theta$ to the end of shared pretraining, apply $\mathcal{M}$, train the subnetwork $f(x; \mathcal{M} \odot \theta_t)$ to completion, and compute accuracy. Prior to composition, we prune each sibling network to the same sparsity $s\%$, and choose a variety of values of $s$ to fill out $\mathcal{M}$'s accuracy-sparsity curve. We compare to baselines of (a) one-shot magnitude pruning with late rewinding \citep{frankle_pruning_2020} and (b) a ``random ticket'' subnetwork with the mask shuffled, but with per-layer pruning ratios maintained.



\begin{figure}[!ht]
\begin{center}
    \includegraphics[width=\textwidth]{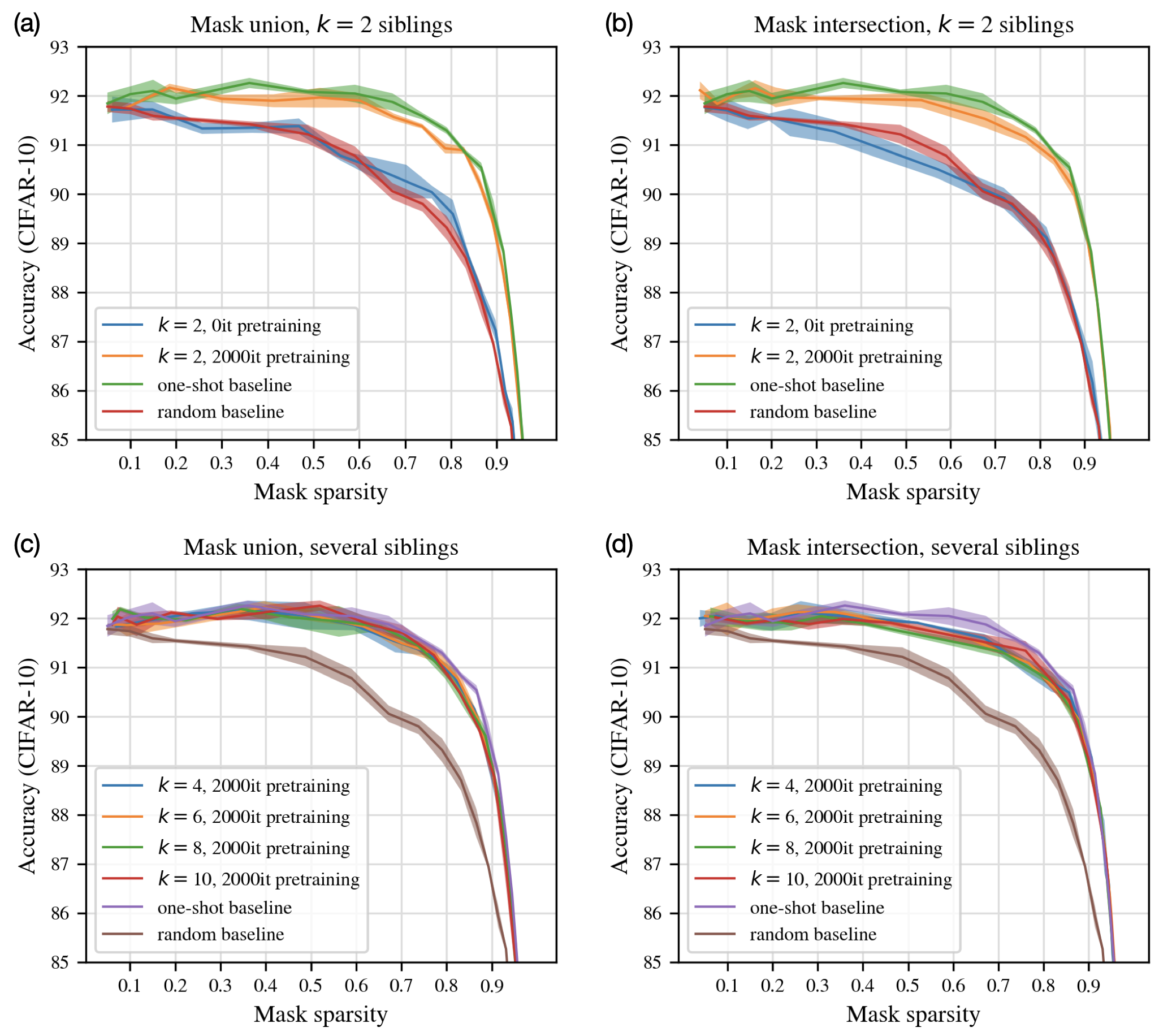}
    \caption{CIFAR-10 accuracy after retraining a ResNet-20 with the composed mask. Mask (a) union and (b) intersection for 2 siblings, Mask (c) union and (d) intersection for 4 to 10 pretrained siblings. For the composed models, the $x$-axis shows the mask sparsity \textit{after} performing union or intersection.}
    \label{fig:compose}
\end{center}
\end{figure}

\textbf{Results.} In Figure \ref{fig:compose}, we plot accuracy vs. sparsity\footnote{Note that the sparsity on the $x$-axis is the post-composition sparsity of $\mathcal{M}$, \textit{not} the per-sibling pruning rate.} of (a) union and (b) intersection masks from $k=2$ siblings, with or without pretraining. Without pretraining, both union and intersection yield masks that are no better than a random pruning baseline (blue and red curves). With pretraining, though, union and intersection masks \textit{both} match the accuracy-sparsity tradeoffs of the one-shot pruning baseline. Further, in parts (c) and (d), we compose up to $k=10$ siblings (pretrained for 2000 iterations) with union and intersection respectively.
Strikingly, several-sibling mask composition still matches the one-shot baseline curve. 

\textbf{Discussion.} We find it particularly surprising that both composition strategies yield almost identical tradeoffs as each other, and with the magnitude pruning baseline. In Section 2, we showed that siblings prune similar weights, motivating the idea of only removing weights when all siblings agree (\textit{i.e.}, taking the mask union). This conservative pruning strategy contrasts with the mask intersection, in which a weight is removed if \textit{any} sibling deems it prunable. But, when final mask sparsity is fixed, our results show that both approaches yield models with the same accuracy, and also match that of single-network magnitude pruning.




\section{Conclusions}

We study the consistency of ResNet-20 masks identified via magnitude pruning. By training sibling networks varying only in their samples of SGD noise, we find that siblings prune more of the same weights than expected by chance. Pruning overlap increases considerably when siblings have undergone shared pretraining, prior to using different samples of SGD noise. Motivated by the significant overlap between siblings' pruned weights, we evaluate whether masks can be combined. We compute the union and intersection of sibling masks, and evaluate these composed masks on CIFAR-10. Surprisingly, both union and intersection achieve very similar performance. Without pretraining, they perform no better than a random ticket, but with pretraining, the combined masks both closely match the accuracy-sparsity tradeoffs of the single-network magnitude pruning baseline. While using the union or intersection does not directly improve performance, mask consistency may offer a potential framework to estimate the confidence of a pruned weight, or to determine optimal iterative pruning ratios. We also wonder whether other methods of composing masks may yield better accuracy-sparsity tradeoffs than a naive union or intersection.

\pagebreak

\bibliography{iclr2021_conference}
\bibliographystyle{iclr2021_conference}

\end{document}